%% file: main.tex
\documentclass[11pt,english]{article}

\usepackage[a4paper,
            bindingoffset=0.2in,
            left=1in,
            right=1in,
            top=1in,
            bottom=1in,
            footskip=.25in]{geometry}

\usepackage{blindtext}\usepackage{graphicx}
\usepackage{amsmath, amssymb}
\usepackage{booktabs}
\usepackage{hyperref}
\usepackage[numbers]{natbib}

\title{A HEART for the environment: Transformer-Based Spatiotemporal Modeling for Air Quality Prediction}
\author{Norbert Bodendorfer \\ Inverence, Madrid, Spain \\ \texttt{norbert.bodendorfer@inverence.com}}
\date{\today}

\begin{document}

\maketitle

\begin{abstract}
Accurate and reliable air pollution forecasting is crucial for effective environmental management and policy-making. 
llull-environment is a sophisticated and scalable forecasting system for air pollution, inspired by previous models currently operational in Madrid and Valladolid (Spain). 
It contains (among other key components) an encoder-decoder convolutional neural network to forecast mean pollution levels for four key pollutants (NO$_2$, O$_3$, PM$_{10}$, PM$_{2.5}$) using historical data, external forecasts, and other contextual features. 
This paper investigates the augmentation of this neural network with an attention mechanism to improve predictive accuracy. 
The proposed attention mechanism pre-processes tensors containing the input features before passing them to the existing mean forecasting model. 
The resulting model is a combination of several architectures and ideas and can be described as a ``Hybrid Enhanced Autoregressive Transformer'', or HEART. 
The effectiveness of the approach is evaluated by comparing the mean square error (MSE) across different attention layouts against the system without such a mechanism. 
We observe a significant reduction in MSE of up to 22\%, with an average of 7.5\% across tested cities and pollutants. The performance of a given attention mechanism turns out to depend on the pollutant, highlighting the differences in their creation and dissipation processes.  
Our findings are not restricted to optimizing air quality prediction models, but are applicable generally to (fixed length) time series forecasting.

\end{abstract}

\section{Introduction}
Air pollution is a major public health and environmental concern, particularly in urban areas where high levels of pollutants pose risks to human health and contribute to climate change \cite{10.3389/fpubh.2020.00014, Gupta2023, Keswani2022}. 
To mitigate these risks, cities start to rely on accurate forecasting models to predict pollution levels and take proactive measures. 

llull-environment \cite{llull-environment-main-paper} is a scalable state-of-the-art air quality forecasting system engineered based on previous experience with deploying such models in Madrid and Valladolid, Spain \cite{DEMEDRANO2021105084}. It provides decision-makers with probabilistic forecasts and activation probabilities for arbitrarily complicated air quality protection protocols. 
It has been shown to provide the currently best available forecasting capabilities of any comparable model \cite{llull-environment-main-paper}. 
A crucial component of this system is an encoder-decoder neural network designed to predict mean pollution levels for four key pollutants: NO$_2$, O$_3$, PM$_{10}$, and PM$_{2.5}$. 
This network utilizes historical pollution data, external forecasts, and other contextual factors to generate predictions.
It relies on spatio-temporal convolutional layers to simultaneously process all measurement stations in a given city following the ideas in \cite{demedrano2020inclusionspatialinformationspatiotemporal}. 

Despite its effectiveness, the current iteration of the model may not fully capture complex temporal dependencies within the input data. Attention mechanisms have demonstrated significant improvements in sequence modeling tasks by enabling models to focus on relevant parts of the input, see \cite{bahdanau2016neuralmachinetranslationjointly, luong-etal-2015-effective, Vaswani2017AttentionIA, Choromanski2020RethinkingAW} for seminal work and \cite{Brauwers_2023} for a review. 
This paper explores the integration of an attention mechanism into the neural network of llull-environment to enhance its predictive performance. The main aim of our work is to evaluate how much an existing forecasting model can be enhanced by adding an attention layer as opposed to developing a new model from scratch. 

We systematically evaluate different attention layer architectures and hyperparameter configurations by measuring the change in mean square error (MSE) compared to the baseline model. The results demonstrate that attention-based enhancements can significantly improve forecast accuracy, with up to a 22\% reduction in MSE observed in some configurations. 

As we will discuss below in more detail, the resulting model is a hybrid of convolutional layers and an attention mechanism as found in contemporary transformers. It is further enhanced with expert-selected autoregressive variables reaching back much further than the input horizon.
As such, it doesn't seem to fit any agreed-upon naming convention. An appropriate nomenclature would be ``Hybrid Enhanced Autoregressive Transformer'', or ``HEART''.

The remainder of this paper is structured as follows.\\
Section \ref{sec:related_work} reviews existing literature on neural networks for air pollution forecasting and attention mechanisms. 
Section \ref{sec:llull-environment-nn} describes the architecture of the baseline model. Section \ref{sec:attention} describes the attention mechanism we are testing and its variations. Section \ref{sec:experiments} outlines the main results and discusses the performance of the attention layers. Finally, we conclude in section \ref{sec:conclusion}.

\section{Related Work} \label{sec:related_work}
Air pollution forecasting has been extensively studied using various statistical and machine learning techniques, see e.g. \cite{bellinger2017, ijerph15040780, Mendez2023, Houdou_2024} for overviews. Traditional approaches, such as autoregressive integrated moving average (ARIMA) models \cite{box1976time}, have been widely used early on, but often struggle with capturing nonlinear dependencies in complex environmental data. More recently, deep learning models, particularly recurrent neural networks (RNNs) and long short-term memory (LSTM) networks \cite{hochreiter1997long, greff2017lstm}, have demonstrated improved performance in time series forecasting tasks and applied to air pollution forecasting, see e.g. \cite{bui2018deeplearningapproachforecasting, su12062570, 8675934, 8784234}. 
%Encoder-decoder architectures have become a common choice for sequence-to-sequence forecasting, enabling effective modeling of temporal dependencies \cite{cho2014learning}. 
However, these architectures may suffer from information bottlenecks when handling long input sequences. To address this, attention mechanisms have been introduced in sequence modeling, allowing neural networks to dynamically weigh different parts of the input sequence \cite{bahdanau2016neuralmachinetranslationjointly, luong-etal-2015-effective, Vaswani2017AttentionIA, Choromanski2020RethinkingAW}. Naturally, such mechanisms have also been tested in air quality prediction tasks, resulting in improved forecasting capabilities \cite{8614140, 9466491, rana2024, pranolo2024}.

%In recent years, the integration of attention mechanisms into neural networks has significantly advanced the modeling of time series data. 
Early on, attention mechanisms have mainly been employed for time series forecasting inside Recurrent Neural Networks as in the seminal paper \cite{bahdanau2016neuralmachinetranslationjointly}. 
Notably, such architectures allow to deal with input and output sequences of variable length, as required by natural language processing where the idea of attention was introduced and is most strongly motivated by heuristic arguments. 
With the advent of transformers \cite{Vaswani2017AttentionIA}, the bottleneck of the recurrent architecture and its inherent problem of poor parallelizability was successfully addressed. Variable length input and output sequences can still be handled by padding (for short sequences) or e.g. by employing copies of the same transformer block connected via a memory mechanism (for sequences longer than the maximal token length). 
An interesting variation of these main lines of thought with a specific focus on time series forecasting can be found in \cite{niu2024attentionrobustrepresentationtime}, see also \cite{kim2024selfattentionseffectivetimeseries} for a review of using attention mechanisms for time series forecasting. 

To finish this section, we briefly discuss the main differences of our model to others in the literature. Since we are using a contemporary attention mechanism containing all the usual features found in modern transformers followed by additional neural network layers, our system is by construction a {\it transformer}. Since it goes beyond simply having a dense layer after the attention mechanism and instead uses a tailored convolutional architecture consisting of several layers, it is a {\it hybrid} of different systems. On top of the typically 72 time steps (hours) of input corresponding to each feature, it uses additional autoregressive features that are given by 72-hour chunks of selected input features at earlier times that are relevant for forecasting during the current horizon. These chunks are selected via the idea of ``social time'', where days are classified not only according to public holidays or similar features, but also via their relation to other holidays and weekends, which influences pollution generation patterns. Hence, the inputs are {\it enhanced} by {\it autoregressive} features selected via expert knowledge, akin to manual implementation of a very long range (up to one year) attention mechanism.  As a consequence of this enhancement, we do not need to consider very long input sequences and the attention mechanism can focus on short range correlations. Since the transformer always receives exactly the maximal token length as input, there is no issue of padding or transferring an internal state via a memory mechanism to another transformer block, leading to superior performance. 

Notable work where related ideas are discussed and implemented includes \cite{gehring2017convolutionalsequencesequencelearning}, where a convolutional architecture is used in conjunction with an attention mechanism for translation tasks.

\section{Brief description of the llull-environment neural network} \label{sec:llull-environment-nn}

In this section, we briefly describe the neural network to be augmented with an attention mechanism in this paper. Since the main content of this paper is neither air pollution forecasting nor the layout of the base network, we will be sparse with details. A comprehensive description is available in \cite{DEMEDRANO2021105084}\footnote{The author does not claim any credit for this architecture and made only minor changes to it.}. We start by outlining two main points that dictate its architecture:

\begin{enumerate}
\item The neural network  is based on the idea of spatial agnosticism for solving spatio-temporal regression problems that was discussed in \cite{demedrano2020inclusionspatialinformationspatiotemporal}. While prior information about the geographical location of a pollution measurement station is in principle available, it is not hardcoded into the neural network a priori. Rather, the neural network is designed to perform spatial convolutions over all measurement stations and to learn correlations between them during training\footnote{We note that there may be relations in the air pollution due to geographical closeness, but other factors such as traffic patterns are likely more important, which is why learning these relations from data makes sense.}.

\item Since the number of measurement stations, the input and output horizons, as well as their temporal relations (typically 72 hours of input followed by 72 hours of output) are known at time of training, it is not necessary to employ recurring structures such as RNNs with LSTM that are designed to deal with series of previously unknown lengths and varying correlations in the data. Rather, we can use a combination of spatio-temporal convolutions and dense layers for greater efficiency.  

\end{enumerate}

Hence, we employ the following encoder-decoder architecture (augmented with suitable activations and regularization mechanisms):
\begin{enumerate}

    \item We start with input tensors arranged in three dimensions $F$, $T$, $S$, for feature, time, and space.  

    \item On each feature, we first perform spatio-temporal convolutions, where we convolve over all measurement stations in the spatial direction and use a certain maximum lag in the time direction. The network uses $H \times S$ distinct convolutions, leading to tensors of dimension $[H, T, S]$ in the latent space after performing a linear combination. The new latent feature set $H$ should be thought of as ``learned'' features combining several input features. 

    \item Next, the learned features from the $H$ dimension are combined by a decoder using convolutions with kernel size $1 \times 1$ due to the expected similarly in combining features across time and space, leading to tensors of dimension $[T, S]$. 

    \item Finally, a dense regressor layer acts on the output and potentially changes the output length if required\footnote{E.g., one could use $168$ hours of input to cover the history of a whole week, while using only 72 hours of output.}. 
    
\end{enumerate}

We note that earlier tests to increase the performance of this layout by simply adding additional dense layers at various points in the architecture did not lead to clear performance increases. Hence, a more targeted approach is likely necessary to further augment the network.

\section{Layout of the attention layers} \label{sec:attention}

\subsection{Attention Mechanisms in Large Language Models} \label{sec:LLMAttention}

\subsubsection{Standard LLM mechanism}

Attention mechanisms \cite{bahdanau2016neuralmachinetranslationjointly, luong-etal-2015-effective, Vaswani2017AttentionIA, Choromanski2020RethinkingAW} have become a cornerstone of modern deep learning models, particularly in the context of large language models (LLMs). They allow a model to dynamically focus on relevant input elements rather than treating all input tokens uniformly. This concept has revolutionized sequence modeling, enabling state-of-the-art performance in various natural language processing tasks.

The key component of attention in LLMs is the self-attention mechanism, which computes attention scores between all elements in an input sequence. This is e.g. formalized as:
\begin{equation}
    \text{Attention}(Q, K, V) = \text{Softmax}\left(\frac{QK^T}{\sqrt{d_k}}\right)V, \label{eq:attention}
\end{equation}
where $Q$, $K$, and $V$ represent the query, key, and value matrices derived from the input sequence, and $d_k$ is a scaling factor to stabilize gradients. The attention scores dictate how much importance should be assigned to different parts of the sequence.

Transformers employ multi-head attention \cite{Vaswani2017AttentionIA} to enhance representation power by learning multiple attention distributions in parallel. This enables the model to capture diverse dependencies within the input data. The introduction of positional encodings \cite{Vaswani2017AttentionIA} ensures that transformers can model sequence order, which is crucial for tasks involving temporal dependencies.
Furthermore, an attention mechanism typically contains dropout layers \cite{srivastava2014dropout}, layer normalization \cite{ba2016layernormalization}, and a residual connection \cite{he2015deepresiduallearningimage} to the original input. 

%In the following sections, we explore how attention mechanisms, inspired by LLMs, can be adapted to enhance air pollution forecasting by capturing intricate temporal relationships in time series data.

\subsubsection{Differences of our attention mechanism to other implementations}

There are two main differences between the llull-environment neural network and an LLM. First, llull-environment operates on input and output sequences of fixed length by a combination of dense and convolutional layers. This makes adding positional encodings unnecessary as the temporal relations are hardcoded in the network architecture. Second, we choose to operate on the univariate time series associated to each feature separately, as opposed to treating it as a multivariate problem (similar to embeddings of words into vector spaces as for LLMs). Hence, we do not need the additional internal dimension of the embedding space and $d_k = 1$ in \eqref{eq:attention}. These two points lead to a few changes in the details of the implementation of the attention mechanism that we will outline below. 

In relation to RNNs, we note that typically, see e.g. \cite{bahdanau2016neuralmachinetranslationjointly}, the attention mechanism is placed between the encoder and the decoder. In transformers, its placement, possibly at multiple stages, varies depending on the model. In our case, we insert the attention mechanism only before the encoder. Here, it can attend to the raw time series data. After the encoder, this data would already have been operated on by spatiotemporal convolutions, likely averaging out and thus hiding some important features. We confirmed this idea in a limited experiment where the T-Att attention (see below) was placed between the encoder and decoder instead, leading to significantly poorer performance than our proposed placement, yet still a measurable improvement over the base model.

\subsection{Application to the llull-environment encoder-decoder architecture}

\subsubsection{General considerations}

We start by noting that our attention mechanism is by construction a mapping from the space of input tensors to itself. As such, it can be thought of as a general non-linear function on the space of input tensors. Our task is hence to find the function that optimally preprocesses input tensors to minimize our chosen loss function (MSE, after applying the remaining parts of the neural network), or rather a network architecture that efficiently learns to approximate that function. 

By standard theorems \cite{cybenko1989approximation}, a given function can be approximated to arbitrary precision using only a single (but very wide) hidden layer along with standard activation functions. 
In practice however, a "smart choice" of network architecture is necessary to achieve an optimal model performance given finite training data and computing resources. 
It must be stated that this is not an exact science and requires a certain amount of experimentation on top of a good intuition. 
In this paper, we apply the lessons learned in constructing attention mechanisms for LLMs as above to time series and compare several architectures. These include a naive ansatz of a combination of several dense layers to check whether an elaborate mechanism including attention weights depending on the input itself are really necessary.

\subsubsection{Tensorial structure and attention}

Input tensors (excluding the batch dimension) have the dimensional structure\footnote{We permuted the indices here as opposed to section \ref{sec:llull-environment-nn}. The reason is that section \ref{sec:llull-environment-nn} focused on spatio-temporal convolutions given a feature $F$, whereas here we operate only on the time direction given a certain feature, but similarly for each measurement station. Hence, the ordering $[S,F,T]$ is most useful for notation.} $[S,F,T]$, where $S$ is the number of measurement stations for a given pollutant, $F$ is the feature dimension, and $T$ is the number of time steps. We typically have $S \approx 5$, $F \approx 30$, and $T = 72$. The choice $T=72$ corresponds to three full days and is owed to the availability of external forecasts at prediction time. 

We expect that an optimal attention mechanism operating on a given univariate time series (for a given station and feature) will depend on the feature, as e.g. a macroscale pollution forecast behaves differently from a categorical calendar variable. 
However, we expect that they will be rather independent of the chosen measurement station, as e.g. a sudden spike in pollution is important independently of the station. 
As a consequence, we train a separate attention mechanism (for temporal dependencies) for each input feature, but operate with it on all stations. 
This ensures increased training data per attention mechanism and reduced computational cost, while the expected loss in model performance is small.

\subsubsection{Optimized attention mechanism}

We now outline the construction of our attention mechanism in detail. 

For each attention head $h$ and input feature $f$, we have three operations $Q^h_f,K^h_f,V^h_f$ from $T$-dimensional vectors (a time series for a given feature / station combination) to $T$-dimensional vectors. They are a concatenation of dense layers and relu activations between the layers, but not at the last layer\footnote{Hence, a layer depth of $1$ corresponds to a linear operation.}. They operate only on the time-indices for their respective feature $f$ in the input tensors, but on each station equally. As a consequence, they preserve the dimension [S,F,T] of an input tensor. In other words, let $s = 1, \ldots, S$ be the station index, $f = 1, \ldots, F$ be the feature index, and let $t = 1, \ldots, T$ be the time index. An input tensor $x_{sft}$ (in index notation) maps as 

\begin{equation} 
x_{sft} \mapsto \left(  M_f^h \ (x_{sf}) \right)_t  ~~ \forall ~~ s, f, h
\end{equation}
where $M \in \{Q, K, V\}$. 

We use $Q$ and $K$ to compute the attention weights $A$ via
\begin{equation} \label{eq:att_weights}
    A(x)_{sft} = \text{Softmax} \left( c^h_f \cdot \, \text{Tanh} \left(Q^h_f(x)_{sft} \cdot K^h_f(x)_{sft} \right) \right), 
\end{equation}
where $c^h_f$ are learnable real parameters and the multiplication of $Q$ and $K$ is component-wise, as opposed to a matrix multiplication in the LLM case which features an additional internal dimension over which can be summed. 
The application of Tanh is for stability purposes: it protects the attention mechanism against input values of previously unobserved magnitude, such as measurement outliers that were not caught by llull-environment's preprocessors, abnormal external forecasts, or rare episodes of extreme pollution. The parameters $c^h_f$ control the maximal relative size of the attention weights due to the boundedness of Tanh.   

After a dropout operation, the attention weights $A(x)$ multiply $V(x)$ component-wise. We now average over the attention heads, apply a layer norm, and finally add the output to the input tensor via a residual connection with learnable parameter. Throughout all operations, suitable dropout layers are applied. We note that there are no projection operations in the different attention heads due to the one-dimensional internal space. 

This design includes all key components of a state-of-the-art attention mechanism, save for the positional encoding that we argued to be unnecessary in our case. It captures strong non-linear dependencies via using stacked layers for $Q, K, V$ as well as influences of all time steps on each other. 
We note that influence of later time steps on earlier ones, as explicitly allowed by our architecture, is not a bug, but a feature: all information contained in the input tensor (e.g. the weather forecast for 3 days ahead) is explicitly known at prediction time and hence will influence human behavior already today. 

In the following, we will describe some variations of this attention mechanism that will be evaluated in this paper. They will test the intuition we developed here. Furthermore, we will test this main attention mechanism for different choices of hyperparameters, namely the number of layers in $Q,K,V$ and the number of attention heads.

\subsubsection{Variations of the attention mechanism for comparative study}

We now outline some variations of the above optimized attention mechanism that we have tested. Each mechanism was initially tested only on a small subset of the  data to evaluate rough performance. The best performing architectures underwent thorough testing for all pollutants in a selection of five different cities. For comparison, we name the above mechanism O-Att for optimized attention. 

\paragraph{O-Att:} The optimized attention mechanism as above. 

\paragraph{Att:} We set $c^h_f$ in \eqref{eq:att_weights} to one in O-Att. This tests whether it is useful to learn the maximal attention weight differences. 

\paragraph{T-Att:} We set $c^h_f = 1$ and make $Q, K, V$ independent of the feature index $f$. This tests wether we need to consider the features individually. 

\paragraph{C-Att:} Same as T-Att, but instead on the time index we operate on the feature index. In other words, the attention mechanism doesn't select important time steps, but important features. 

\paragraph{TC-Att:} A concatenation of T-Att followed by C-Att.

\paragraph{M-Att:} (Mock attention). Same as Att, but $Q$ and $K$ are not computed and $A_{sft}=1$ component-wise. This tests whether it is necessary to compute attention weights $A$ via $Q$ and $K$ or whether $V$ along with the residual connection is already performant enough. 

\paragraph{NL-Att:} Similar to Att, but the univariate time series is first embedded into a higher-dimensional space (similar to the embedding space for usual LLMs) via a learnable mapping. As a consequence, the attention weights are $T \times T$ matrices for each $s,f$ and are hence ``non-local'' in the time direction.

\paragraph{Dense:} The attention mechanism is simply a stacking of dense layers and relu activations for each feature. There are no operations $Q,K,V$, no layer norm, no residual connection. There is only a final dropout layer and a sum over attention heads afterwards. This tests wether we can not simply use a stacking of dense layers.

\section{Experiments and Results} \label{sec:experiments}
% Description of experiments, datasets, and results analysis

\subsection{Input data and training}

We perform experiments on historical pollution data from five Spanish cities, see table \ref{tab:pollutant_stations}. The data goes back to January 2013. Training data starts at the beginning of 2014 due to the construction of auto-regressive input data up to one year back. We use the first 90\% of the data for training and the last 10\% as a validation set. Training is via the Adam optimizer with a progressively decreasing learning rate. We decrease the learning rate (and eventually stop) once the validation MSE is not decreasing any more for a certain number of epochs. We use the lowest achieved MSE of the validation data as a measurement. 

\begin{table}[h]
    \centering
    \begin{tabular}{lcccc}
        \toprule
        City / stations      & NO$_2$ & O$_3$ & PM$_{10}$ & PM$_{2.5}$ \\
        \midrule
        Gijón      & 6  & 5  & 6  & 5  \\
        Granada    & 2  & 1  & 2  & 2  \\
        Málaga     & 5  & 4  & 4  & 2  \\
        Valencia   & 7  & 6  & 5  & 5  \\
        Valladolid & 5  & 3  & 4  & 4  \\
        \bottomrule
    \end{tabular}
    \caption{Number of measurement stations for each pollutant in different cities}
    \label{tab:pollutant_stations}
\end{table}

The pollution data is freely available from the European Environment Agency (EEA) \cite{eea2024}. It has undergone cleaning, outlier handling, and imputation using proprietary algorithms of llull-environment. Additional features include auto-regressive variables using ``social time'' (similar days of the last weeks and year), macroscale pollution forecasts (via the Copernicus Atmosphere Monitoring Service (CAMS)) \cite{cams2024}, weather forecasts from the Weather Research \& Forecasting Model (WRF) \cite{wrf_model}, and calendar data.

\subsection{Broad comparison}

\begin{figure}
    \centering
    \includegraphics[width=1\linewidth]{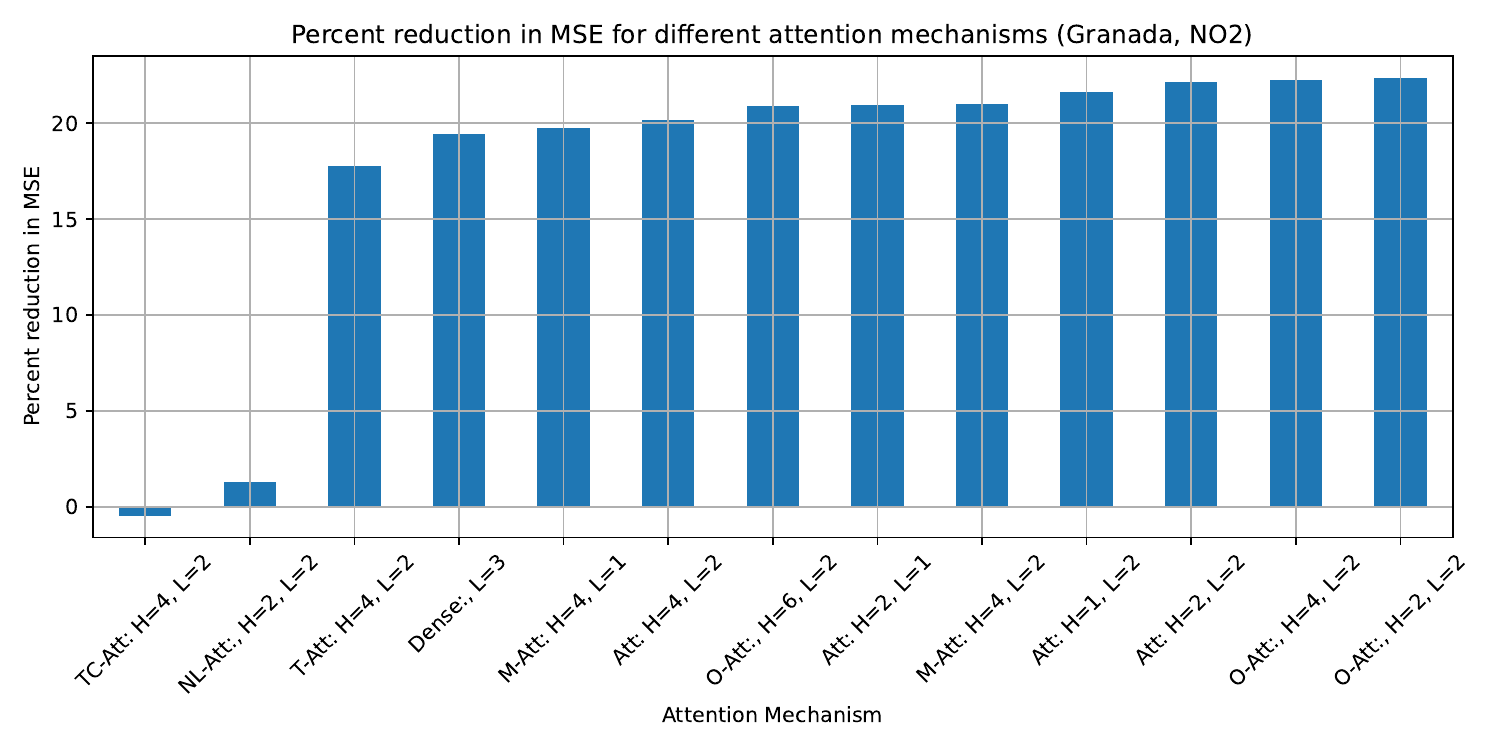}
    \caption{Percentage reduction in MSE as opposed to no attention mechanism for Granada and pollutant NO2. H indicates the number of attention heads, L the number of layers in $Q, K, V$ or dense layers respectively. }
    \label{fig:Granada_NO2}
\end{figure}

\begin{figure}
    \centering
    \includegraphics[width=1\linewidth]{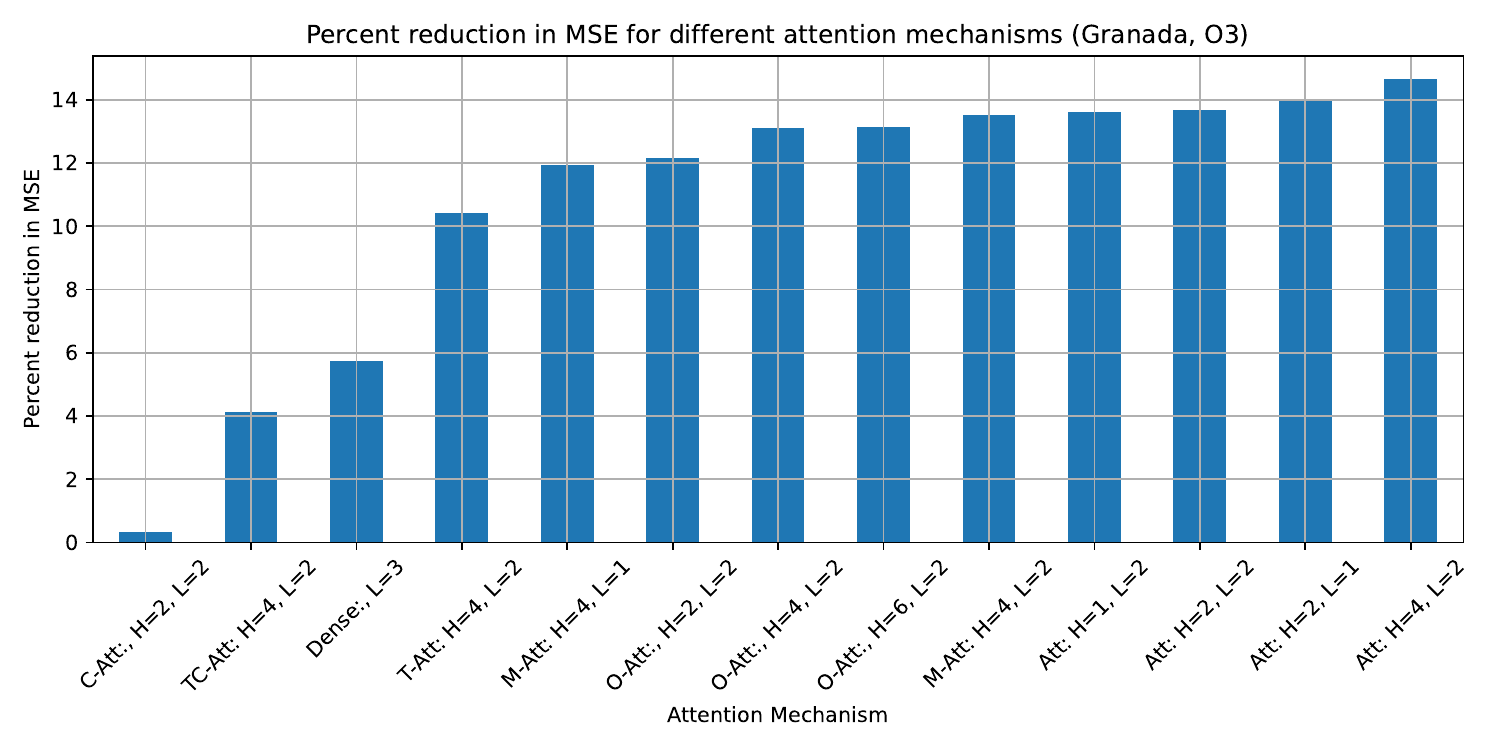}
    \caption{Percentage reduction in MSE as opposed to no attention mechanism for Granada and pollutant O3. H indicates the number of attention heads, L the number of layers in $Q, K, V$ or dense layers respectively. }
    \label{fig:Granada_O3}
\end{figure}

We first apply all algorithms to NO2 and O3 forecasts in Granada. The choice is due to the reduced training time due to the small amount of measurement stations available. Our results are summarized in figure \ref{fig:Granada_NO2} and \ref{fig:Granada_O3}.

We observe that Att, O-Att, and M-Att are most performant. It is unlikely that the inferior performance of the other architectures was due to statistical fluctuations during training. 
Hence, we focus on Att, O-Att, and M-Att in the following. 

These results are not surprising: it is intuitively clear that attention mechanisms such as T-Att have only a limited capacity due to treating all input features in the same footing. 
Similarly, using only dense layers without a residual connection should be expected to be less performant. 
The additional higher-dimensional embedding in NL-Att should also not be expected to lead to increased performance (but the increase in training time was significant).

\subsection{Detailed comparison}

\begin{figure}
    \centering
    \includegraphics[width=1\linewidth]{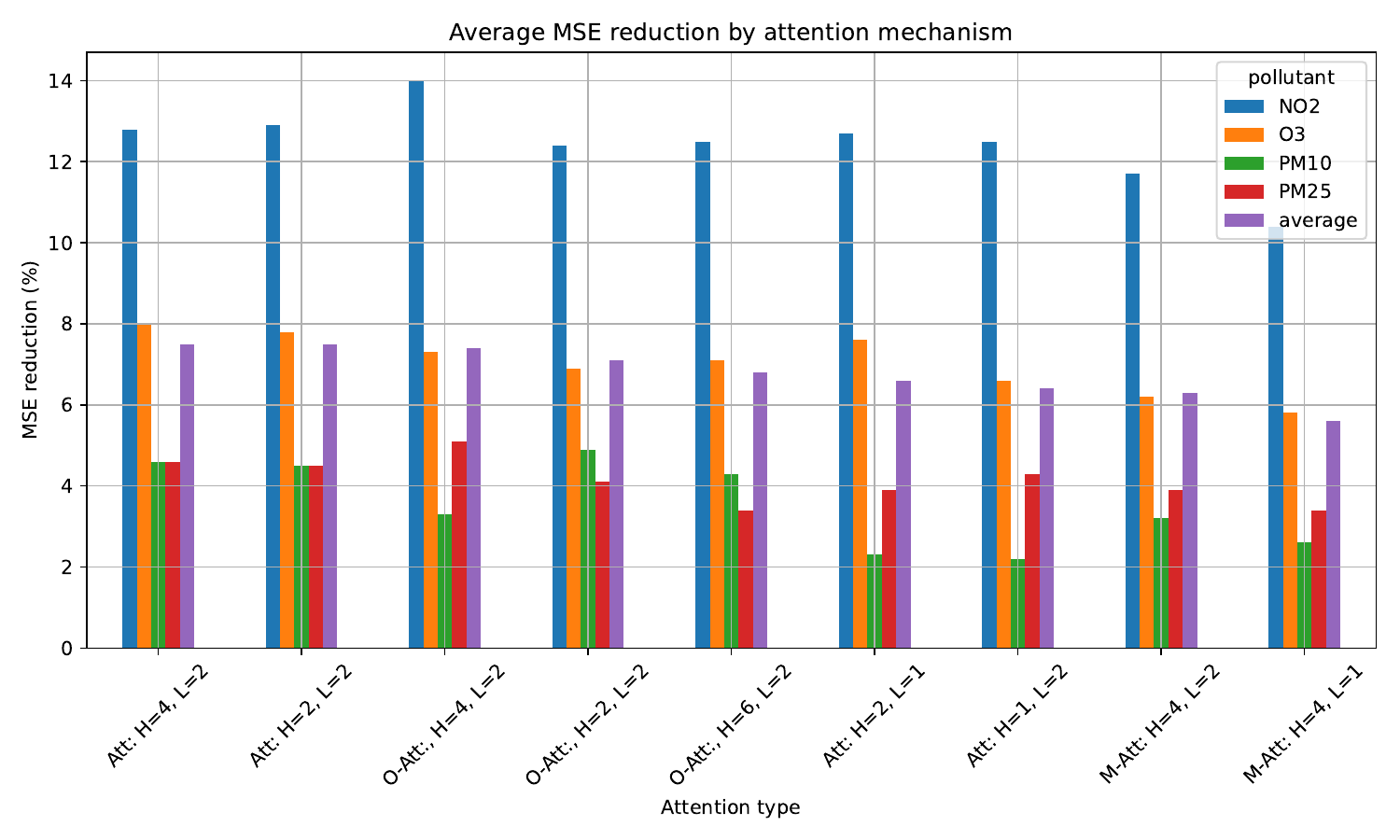}
    \caption{Reduction of MSE averaged over cities, sorted by pollution average.}
    \label{fig:average_figure}
\end{figure}

In the detailed comparison, we train the model for all four pollutants in each of the five cities. Detailed results for individual cities are presented in the tables \ref{tab:MSEGainByCityGijon}, \ref{tab:MSEGainByCityGranada}, \ref{tab:MSEGainByCityMalaga}, \ref{tab:MSEGainByCityValencia}, \ref{tab:MSEGainByCityValladolid}. Averages over the cities are shown in table \ref{tab:MSEGainAverage} and vizualized in figure \ref{fig:average_figure}. 

We make the following observations
\begin{itemize}
    \item We observe a different level of performance gain depending on the pollutant. This is not surprising as their creation and dissipation mechanisms significantly differ and hence attention across the time series is expected to have different levels of potential benefits. 
    \item The relative performance of the different attention mechanisms varies slightly from city to city. This is not surprising as there are typical statistical fluctuations of a few percent in the validation MSE due to randomness during the training. This is reflected in similar fluctuations in the percentage reduction of MSE shown in the tables. Hence, table \ref{tab:MSEGainAverage} containing the averages over all five cities is most relevant for judging the different attention mechanisms.  
    \item On average, Att with two layers and 2 or 4 heads seems to have the best performance. It is not surprising that Att performs better than M-Att, as the latter lacks attention weights computed by the key and query matrices. 
    \item It is somewhat surprising that O-Att doesn't perform as well as Att, even though it is more expressive in that the threshold of attention differences can be learned. It could be that this additional freedom is not beneficial as it may lead to instabilities due to rapid changes in the attention weights during training. Only repeated training and averaging over the results will likely be able to decide this issue.  
    \item In general, the difference between the attention mechanisms seems statistically significant, but not very large. Hence, e.g. M-Att with 2 layers could be chosen over Att if there were significant computational constraints associated.

\end{itemize}

\input{tables_all_cities}

\input{tables_averages}

\newpage

\section{Conclusion} \label{sec:conclusion}
% Insights and interpretation of results

In this paper, we have tested various attention mechanisms to augment the endocer-decoder neural network of llull-environment for air pollution forecasting. We found consistent improvement in MSE due to the added attention layer. 
Our work adds evidence to the well-known observation that including attention layers to existing neural networks can significantly boost performance. 
By leaving the previous architecture unchanged, it is relatively simple to test such an augmentation in other existing systems and our work encourages to generally do so.  
If every percent of performance counts in a given application, one would have to rerun training several times and use techniques such as k-fold cross validation to better gauge the exact gains. 
If done so in the present context, it would not be surprising to find that different pollutants are optimally handled by different attention layouts. 
This however goes beyond the scope of this paper. 

An interesting observation is that the encoder-decoder architecture described in section \ref{sec:llull-environment-nn} heavily relies on all of the input features having the same time horizon and segmentation, i.e. corresponding to values measured in or forecasted for a certain shared set of time steps. Prepending an attention mechanism as described here that operates separately on each feature allows to drop that requirement, as the attention layer functions as a preprocessor for the features and learns to automatically align them to the same time steps before the encoder-decoder block operates. This makes the architecture more flexible: one can for example add features in arbitrarily shifted time horizons (such as autoregressive inputs) or use longer time horizons with data that is measured less frequently.

\section*{Acknowledgments}

The author would like to thank Victor de Buen Remiro for frequent discussions around air quality modeling and many fruitful interactions in the llull-environment project. 

\bibliographystyle{IEEEtran}
%\AtEveryBibitem{\clearfield{doi} \clearfield{url} \clearfield{issn}}
\bibliography{references} % You will need a references.bib file

\end{document}

%% file: tables_all_cities.tex
\begin{table}
\begin{center}
\begin{tabular}{lccccc}
\toprule
pollutant & NO2 & O3 & PM10 & PM25 & average \\
attention &  &  &  &  &  \\
\midrule
O-Att:, H=4, L=2 & \textbf{8.9} & 5.9 & 4.9 & \textbf{9.2} & \textbf{7.2} \\
Att: H=2, L=1 & 6.2 & \textbf{7.1} & \textbf{5.5} & 8.7 & 6.9 \\
Att: H=4, L=2 & 8.8 & 6.2 & 4.3 & 7.9 & 6.8 \\
O-Att:, H=6, L=2 & 8.5 & 5.8 & 4.6 & 7.7 & 6.6 \\
O-Att:, H=2, L=2 & 8.3 & 4.5 & 4.7 & 7.7 & 6.3 \\
Att: H=2, L=2 & 8.2 & 4.9 & 4.1 & 7.5 & 6.2 \\
Att: H=1, L=2 & 7.5 & 4.2 & 4.5 & 8.2 & 6.1 \\
M-Att: H=4, L=2 & 6.9 & 4.6 & 3.8 & 8.0 & 5.8 \\
M-Att: H=4, L=1 & 4.1 & 5.0 & 3.3 & 7.6 & 5.0 \\
\bottomrule
\end{tabular}
\end{center}
\caption{Percent reduction in MSE for different attention mechanisms for Gijón.}
\label{tab:MSEGainByCityGijon}
\end{table}

\begin{table}
\begin{center}
\begin{tabular}{lccccc}
\toprule
pollutant & NO2 & O3 & PM10 & PM25 & average \\
attention &  &  &  &  &  \\
\midrule
O-Att:, H=2, L=2 & \textbf{22.4} & 12.2 & \textbf{3.7} & 3.3 & \textbf{10.4} \\
Att: H=2, L=2 & 22.2 & 13.7 & 1.6 & 2.6 & 10.0 \\
O-Att:, H=4, L=2 & 22.3 & 13.1 & -0.1 & 4.5 & 9.9 \\
Att: H=4, L=2 & 20.2 & \textbf{14.6} & 0.0 & 1.4 & 9.1 \\
Att: H=1, L=2 & 21.6 & 13.6 & -7.0 & \textbf{4.8} & 8.3 \\
M-Att: H=4, L=2 & 21.0 & 13.5 & -3.3 & -0.5 & 7.7 \\
O-Att:, H=6, L=2 & 20.9 & 13.1 & -0.0 & -3.7 & 7.6 \\
M-Att: H=4, L=1 & 19.8 & 11.9 & -2.5 & -0.3 & 7.2 \\
Att: H=2, L=1 & 20.9 & 14.0 & -6.6 & -0.2 & 7.0 \\
\bottomrule
\end{tabular}
\end{center}
\caption{Percent reduction in MSE for different attention mechanisms for Granada.}
\label{tab:MSEGainByCityGranada}
\end{table}

\begin{table}
\begin{center}
\begin{tabular}{lccccc}
\toprule
pollutant & NO2 & O3 & PM10 & PM25 & average \\
attention &  &  &  &  &  \\
\midrule
O-Att:, H=6, L=2 & \textbf{14.1} & 10.3 & \textbf{12.7} & 2.5 & \textbf{9.9} \\
Att: H=4, L=2 & 13.2 & 11.4 & 10.4 & \textbf{2.6} & 9.4 \\
O-Att:, H=2, L=2 & 12.8 & 10.9 & 10.9 & 0.5 & 8.8 \\
Att: H=2, L=2 & 12.9 & 11.2 & 10.9 & 0.3 & 8.8 \\
O-Att:, H=4, L=2 & 13.6 & \textbf{11.9} & 9.3 & 0.1 & 8.7 \\
M-Att: H=4, L=2 & 11.8 & 9.8 & 7.4 & 1.8 & 7.7 \\
Att: H=2, L=1 & 13.1 & 9.2 & 6.1 & 0.8 & 7.3 \\
Att: H=1, L=2 & 12.2 & 9.3 & 5.4 & 1.9 & 7.2 \\
M-Att: H=4, L=1 & 10.0 & 7.9 & 5.5 & 2.4 & 6.5 \\
\bottomrule
\end{tabular}
\end{center}
\caption{Percent reduction in MSE for different attention mechanisms for Málaga.}
\label{tab:MSEGainByCityMalaga}
\end{table}

\begin{table}
\begin{center}
\begin{tabular}{lccccc}
\toprule
pollutant & NO2 & O3 & PM10 & PM25 & average \\
attention &  &  &  &  &  \\
\midrule
Att: H=2, L=1 & \textbf{16.5} & \textbf{4.0} & 1.6 & \textbf{7.4} & \textbf{7.4} \\
Att: H=4, L=2 & 14.4 & 2.8 & 1.2 & 6.1 & 6.2 \\
Att: H=2, L=2 & 14.5 & 3.6 & -0.1 & 6.4 & 6.1 \\
O-Att:, H=4, L=2 & 16.2 & 2.1 & -0.7 & 6.6 & 6.1 \\
M-Att: H=4, L=1 & 10.8 & 1.1 & \textbf{2.0} & 6.6 & 5.1 \\
O-Att:, H=6, L=2 & 11.6 & 1.9 & 1.6 & 5.0 & 5.0 \\
O-Att:, H=2, L=2 & 13.6 & 2.6 & -1.4 & 4.4 & 4.8 \\
Att: H=1, L=2 & 14.0 & 1.9 & -0.6 & 2.8 & 4.5 \\
M-Att: H=4, L=2 & 12.0 & -0.4 & -0.4 & 2.1 & 3.3 \\
\bottomrule
\end{tabular}
\end{center}
\caption{Percent reduction in MSE for different attention mechanisms for Valencia.}
\label{tab:MSEGainByCityValencia}
\end{table}

\begin{table}
\begin{center}
\begin{tabular}{lccccc}
\toprule
pollutant & NO2 & O3 & PM10 & PM25 & average \\
attention &  &  &  &  &  \\
\midrule
M-Att: H=4, L=2 & 7.0 & 3.7 & 8.4 & \textbf{7.8} & \textbf{6.7} \\
Att: H=2, L=2 & 6.9 & \textbf{5.6} & 6.2 & 5.9 & 6.2 \\
Att: H=4, L=2 & 7.4 & 5.0 & 7.1 & 4.8 & 6.1 \\
Att: H=1, L=2 & 7.3 & 3.8 & \textbf{8.9} & 3.6 & 5.9 \\
O-Att:, H=4, L=2 & \textbf{9.0} & 3.7 & 3.3 & 4.9 & 5.2 \\
O-Att:, H=6, L=2 & 7.5 & 4.5 & 2.6 & 5.7 & 5.1 \\
O-Att:, H=2, L=2 & 4.8 & 4.1 & 6.5 & 4.5 & 5.0 \\
Att: H=2, L=1 & 7.0 & 3.7 & 4.8 & 2.8 & 4.6 \\
M-Att: H=4, L=1 & 7.6 & 3.1 & 4.8 & 0.6 & 4.0 \\
\bottomrule
\end{tabular}
\end{center}
\caption{Percent reduction in MSE for different attention mechanisms for Valladolid.}
\label{tab:MSEGainByCityValladolid}
\end{table}

%% file: tables_averages.tex
\begin{table}
\begin{center}
\begin{tabular}{lccccc}
\toprule
pollutant & NO2 & O3 & PM10 & PM25 & average \\
attention &  &  &  &  &  \\
\midrule
Att: H=4, L=2 & 12.8 & \textbf{8.0} & 4.6 & 4.6 & \textbf{7.5} \\
Att: H=2, L=2 & 12.9 & 7.8 & 4.5 & 4.5 & \textbf{7.5} \\
O-Att:, H=4, L=2 & \textbf{14.0} & 7.3 & 3.3 & \textbf{5.1} & 7.4 \\
O-Att:, H=2, L=2 & 12.4 & 6.9 & \textbf{4.9} & 4.1 & 7.1 \\
O-Att:, H=6, L=2 & 12.5 & 7.1 & 4.3 & 3.4 & 6.8 \\
Att: H=2, L=1 & 12.7 & 7.6 & 2.3 & 3.9 & 6.6 \\
Att: H=1, L=2 & 12.5 & 6.6 & 2.2 & 4.3 & 6.4 \\
M-Att: H=4, L=2 & 11.7 & 6.2 & 3.2 & 3.9 & 6.3 \\
M-Att: H=4, L=1 & 10.4 & 5.8 & 2.6 & 3.4 & 5.6 \\
\bottomrule
\end{tabular}
\end{center}
\caption{Percent reduction in MSE for different attention mechanisms, average over cities.}
\label{tab:MSEGainAverage}
\end{table}